%% file: main.tex
\documentclass[10pt,twocolumn,letterpaper]{article}

\usepackage[algorithms]{wacv}

\usepackage{wrapfig}
\usepackage{multirow}
\usepackage{comment}
\usepackage{amsmath}

\definecolor{wacvblue}{rgb}{0.21,0.49,0.74}
\usepackage[pagebackref,breaklinks,colorlinks,allcolors=wacvblue]{hyperref}

\def\wacvPaperID{1757} 
\def\confName{WACV}
\def\confYear{2026}

\title{Exploring Compositionality in Vision Transformers using Wavelet Representations}

\author{
Akshad Shyam Purushottamdas \\
\texttt{ai22mtech02006@iith.ac.in}
\and
Pranav K Nayak \\
\texttt{pranav.kalsanka@alumni.iith.ac.in}
\and
Divya Mehul Rajparia \\
\texttt{es22btech11013@iith.ac.in}
\and
Deekshith Patel \\
\texttt{ai23btech11003@iith.ac.in}
\and
Yashmitha Gogineni \\
\texttt{es22btech11014@iith.ac.in}
\and
Konda Reddy Mopuri \\
\texttt{krmopuri@ai.iith.ac.in}
\and
Sumohana S. Channappayya \\
\texttt{sumohana@ee.iith.ac.in} \\
\\
IIT Hyderabad, India
}

\begin{document}
\maketitle
\begin{abstract}
While insights into the workings of the transformer model have largely emerged by analysing their behaviour on language tasks, this work investigates the representations learnt by the Vision Transformer (ViT) encoder through the lens of compositionality. We introduce a framework - analogous to the compositionality setting proposed for representation learning in~\cite{andreas2019measuring} - to test for compositionality in the ViT encoder. Crucial to drawing this analogy is the Discrete Wavelet Transform (DWT), which is a simple yet effective tool for getting input-dependent primitives in the vision setting. By examining the ability of composed representations to reproduce original image representations, we empirically test the extent to which compositionality is respected in the representation space. Our findings show that primitives from a one-level DWT decomposition produce encoder representations that approximately compose in latent space, offering a new perspective on how ViTs structure information.

\end{abstract}

\section{Introduction}
\label{sec:introdution}
    Vision Transformers (ViTs), in their supervised~\citep{dosovitskiy2020image}, self-supervised~\citep{caron2021emerging}, and unsupervised~\citep{he2022masked} variants, have delivered state-of-the-art performance across various computer vision applications. Image classification~\citep{dosovitskiy2020image}, object detection~\citep{li2022exploring}, semantic segmentation~\citep{strudel2021segmenter}, and image captioning and generation~\citep{radford2021learning} are a few examples. ViTs leverage the transformer architecture - originally popularized in natural language processing (NLP) tasks - to process images using self-attention. This utilisation of transformer architecture in computer vision has opened new avenues for understanding and processing visual data. 



It is natural to wonder why ViTs deliver such performance despite their origins in language models. Given their prevalence as backbones for generating image embeddings for various downstream tasks, we focus our investigation on these embeddings themselves. Several works have investigated the inner workings of the ViT. \citep{raghu2022vision} show that the representations of ViT encoder layers are much more uniform than the CNN-based architectures. \citep{park2022vision} sheds light on the  Multi-head Self Attention block and its optimization. \citep{bhojanapalli2021understanding} test the ViT's robustness to input and model perturbations. Their correlation analysis led to interesting findings about ViT models organizing themselves into correlated groups.
Our motivation is along the lines of such studies attempting to understand the representations learned by ViTs and make them more explainable. The main contributions of this paper are summarized as follows.
\begin{enumerate}
    \item A framework for testing compositionality in ViT encoder representations, analogous to the framework proposed by~\cite{andreas2019measuring} for representation learning.
    \item The use of the Discrete Wavelet Transform (DWT) to generate basis sets (input-specific primitives) for images. To the best of our knowledge, previous works have not used this approach to analyse ViTs.
    \item Promising empirical results that demonstrate compositionality in the encoder representations of the ViT. Our analysis reveals that ViT patch representations at the final encoder layer are compositional for the DWT primitives obtained by a one-level decomposition.
\end{enumerate}



\section{Background}
\label{sec:background}
\subsection{Vision Transformers}
Following the success of transformers \cite{vaswani2023attention} in NLP, \citep{dosovitskiy2020image} adapted them to vision tasks. ViTs divide an input image into patches, each of which is tokenized. Positional embeddings are added to each token embedding to preserve its spatial location. A special CLS token is appended to the input embeddings, and is used for the final classification. The dimension of all the patch representations remains constant throughout the encoder layers, which gives the ViT model flexibility. 

\subsection{Compositionality in Representation Learning}
\label{subsec:comp_in_rep}
Representational compositionality has been a field of study since the days of the connectionist approach~\citep{FODOR19883, Chalmers1990WhyFA}. Its linguistic origins still make themselves known in current research, with most investigations focusing on representations in NLP tasks and models~\citep{Chen2023SkillsinContextPU, Li2023DissectingCC, Dziri2023FaithAF}.~\citep{janfreg2001} defines the principle of compositionality as ``the meaning of a compound expression is a function of the meaning of its parts and of the syntactic rule by which they are combined". The notions of \textit{meaning} and \textit{syntactic rules} in language model representations naturally lend themselves to the study of compositionality.

Formally, a compositional representation function learns a homomorphism between the input space and the its representation space ~\citep{andreas2019measuring}. A homomorphism $\phi: H \rightarrow G$ is a map between two groups $(H, \cdot)$ and $(G, \oplus)$, such that if $\phi(h_1)=g_1$ and $\phi(h_2) = g_2$ for $h_1, h_2 \in H$ and $g_1, g_2 \in G$, then $\phi(h_1 \cdot h_2) = g_1 \oplus g_2$.

The study of compositional nature of pretrained models is motivated, in part, by interpretability. A model that can break its input into meaningful pieces and reconstruct it in a human-understandable manner is more interpretable than a model that does not. With interpretability in mind, we pursue our investigations into the representations learned by ViT. 

In the NLP domain, such investigations usually decompose the input space into a dictionary of words, which acts as the fundamental set used to represent all sentences. Note that each word in the dictionary (the fundamental set) has a standalone meaning, independent of the sentence it is used in. However, it is difficult to construct a dictionary of {\em visually meaningful} images in the image domain, since the image space is continuous. In other words, we cannot construct a dictionary with infinite cardinality. This difficulty is additionally compounded by the uninterpretable nature of the canonical basis in the image space - the set of $H \times W \times C$ matrices with every element being 0 except for a single 1 at some position. Thus, we propose a different approach to decompose an image into its visually meaningful primitives, turning to analytical tools from signal processing.

\subsection{Discrete Wavelet Transform (DWT)}

\begin{figure*}[t!]
    \centering

    \begin{subfigure}{0.22\textwidth}
        \centering
        \includegraphics[width=\textwidth]{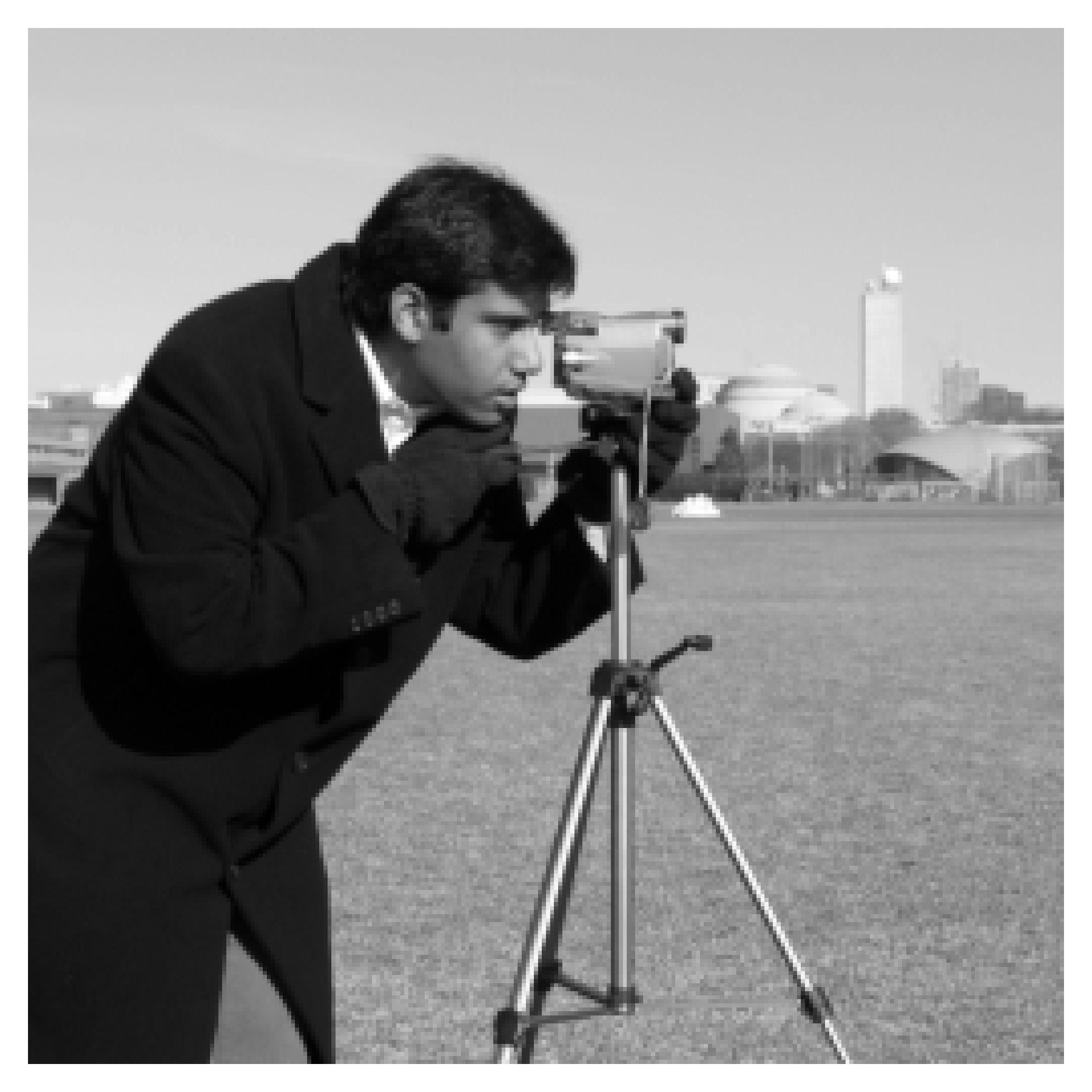}
        \caption{Original image}
    \end{subfigure}
    \hfill
    \begin{subfigure}{0.22\textwidth}
        \centering
        \includegraphics[width=\textwidth]{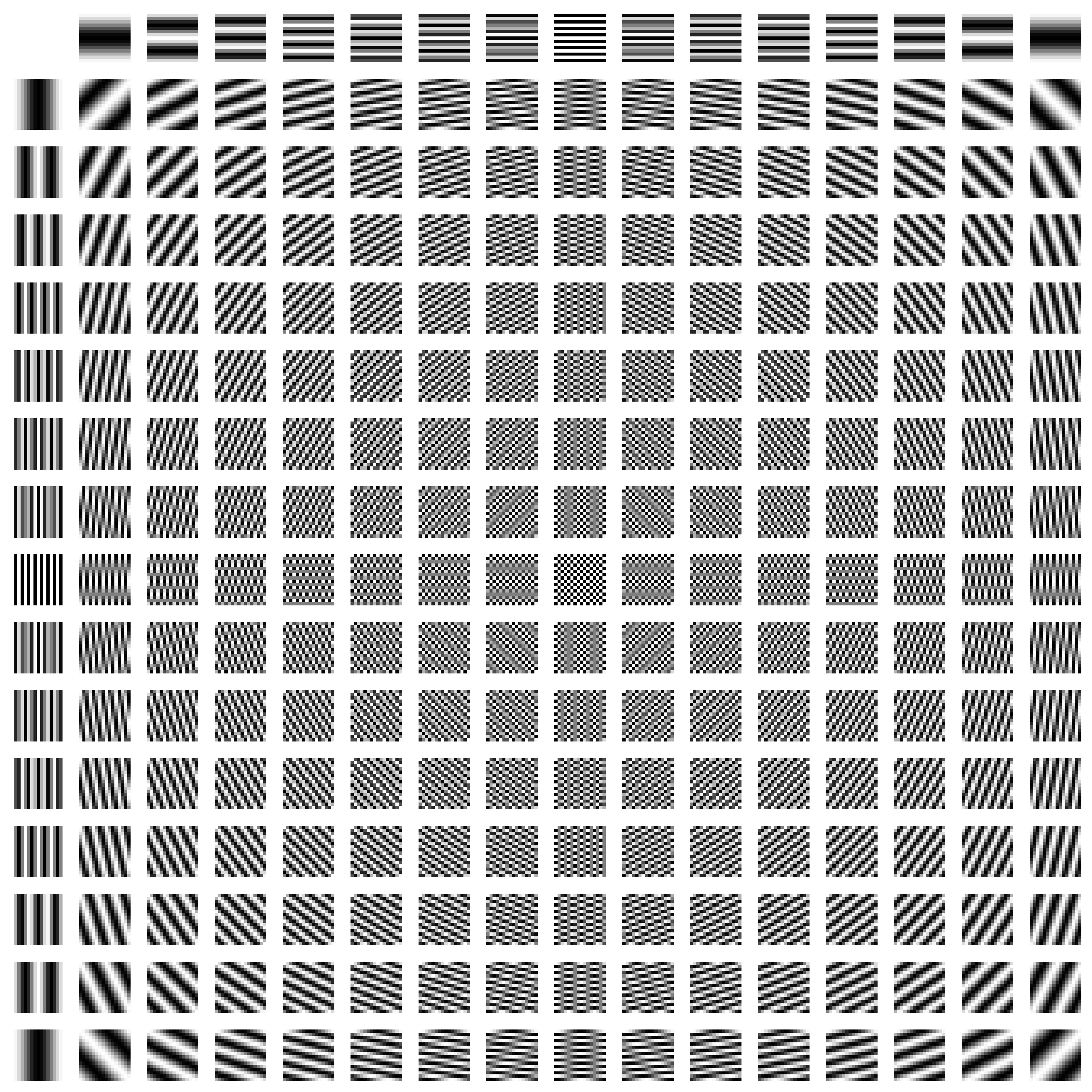}
        \caption{DFT basis}
    \end{subfigure}
    \hfill
    \begin{subfigure}{0.22\textwidth}
        \centering
        \includegraphics[width=\textwidth]{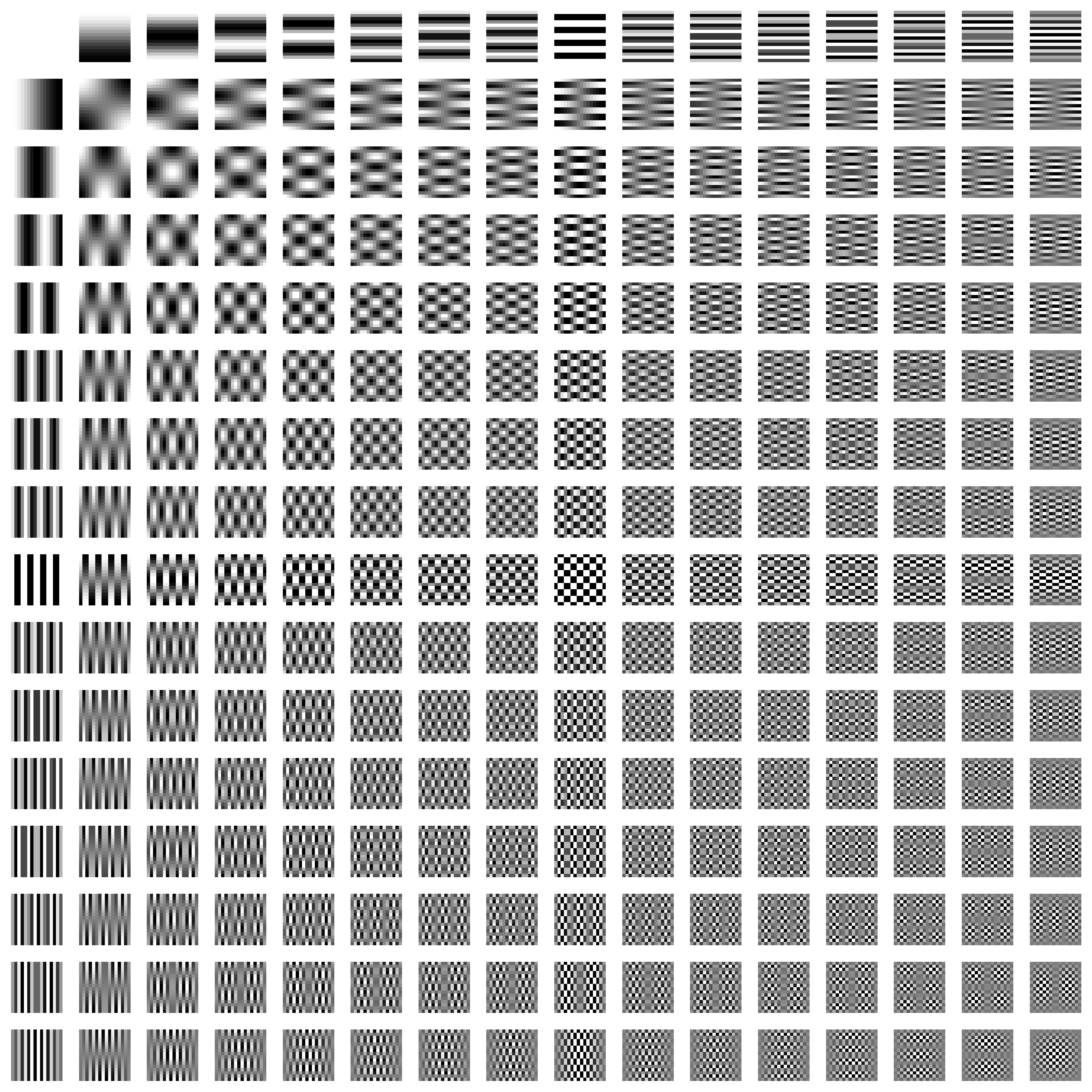}
        \caption{DCT basis}
    \end{subfigure}
    \hfill
    \begin{subfigure}{0.22\textwidth}
        \centering
        \includegraphics[width=\textwidth]{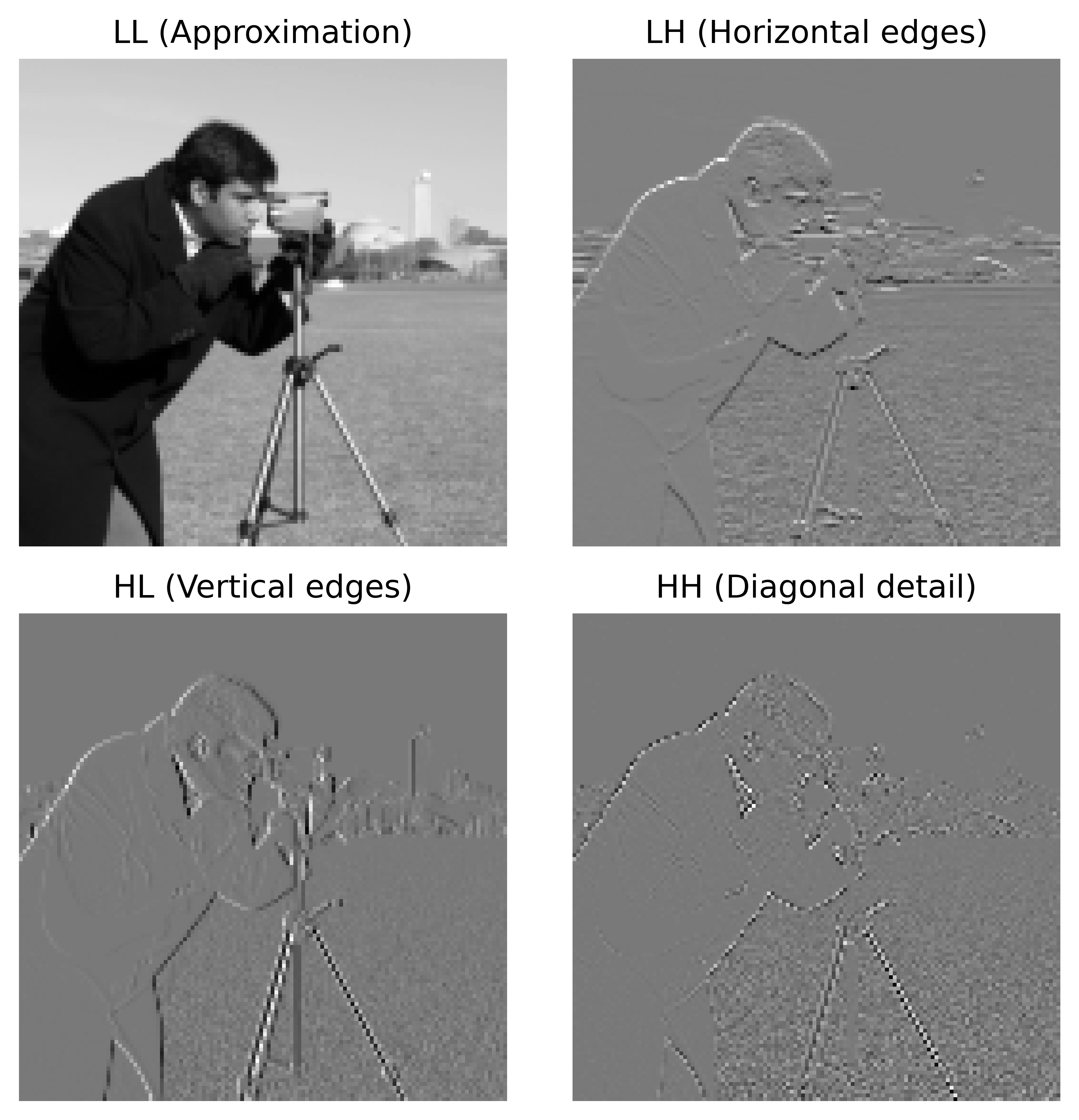}
        \caption{DWT subbands}
    \end{subfigure}

    \caption{
    Comparison of DFT, DCT, and DWT decompositions. Unlike DFT and DCT, whose subbands represent global frequency components, DWT produces spatially localized, visually interpretable subbands suitable for compositionality analysis.
    }
    \label{fig:frequency_vs_wavelet}
\end{figure*}

While the Fourier series and Fourier transform are excellent tools to analyze the frequency spectrum of images, they do not provide localization in the pixel domain. Simply put, the Fourier spectrum of an image is not visually meaningful.

Both the Discrete Fourier Transform (DFT) and the Discrete Cosine Transform (DCT) decompose an image into a set of basis functions; however, these bases themselves lack semantic or visual interpretability. In particular, the global sinusoidal and cosine bases used by the DFT and DCT do not correspond to localized or meaningful image structures, analogous to how individual words in a sentence carry standalone meaning. As a result, although such transforms are effective for energy compaction and frequency analysis, their coefficients are difficult to interpret in terms of image content or structure, limiting their suitability for compositional analysis.

The DWT~\cite{daubechies1992ten} stands out among time-frequency analysis tools due to its unique ability of time-frequency localization. Specifically, applying the DWT to an image decomposes it into sub-bands that correspond to localized spatial-frequency components (e.g., horizontal, vertical, and diagonal details at multiple scales), which are visually interpretable and content-aware primitives. The invertibility of the sub-band decomposition enables lossless reconstruction, making the DWT our tool of choice for compositionality analysis. 

Formally, given any $W\times H$ image $I$, it can be represented in terms of its wavelet coefficients as,
\begin{equation}
\begin{split}
    I_{W\times H} &= \sum\limits_{i=0}^{W-1}\sum\limits_{j=0}^{H-1}A_{M,i,j}\phi_{M,i,j}\\ &+ \sum\limits_{m=1}^{M}\sum\limits_{i=0,j=0}^{W-1,H-1}\sum\limits_{k=1}^{3}D_{m,i,j}^{k}\psi_{m,i,j}^{k}, 
\end{split}
\label{eq:sum}
\end{equation} 

where $A_{M,i,j} = \langle I_{W \times H},\phi_{M,i,j}\rangle$ and $D_{m,i,j}^{k} = \langle I_{W \times H},\psi_{m,i,j}^{k} \rangle$  are the approximation and detail coefficients respectively. $k$ is the sub-band index, and the functions $\phi$ and $\psi$ are the scaling (approximation) and wavelet (detail) wavelet bases. An orthogonal decomposition is assumed in this work. The first term corresponds to the approximation of the image at level $M$, while the second term represents all of the detail coefficients from level 1 to $M$. 

It is important to emphasize that, unlike words in a sentence whose meanings exist independently of the sentence itself, the resulting sub-band coefficients are inherently dependent on the specific input image. Hence, these are {\em input-dependent primitives}.

After the introduction of ViTs, the DWT has been used for lossless downsampling to address their efficiency-vs-accuracy tradeoff \citep{yao2022wavevitunifyingwavelettransformers}. \cite{zhang2024waveletformernettransformerbasedwaveletnetwork} also employs the DWT to improve the quality of the input in a transformer-based network. However, to our knowledge, the DWT has not been used to explore compositionality in ViTs.

\subsection{Compositionality of Image Representations}


When inputs belong to the pixel space, and a neural network learns input representations, the groups across which compositionality is studied are vector spaces. These spaces need to be equipped with a binary operation that satisfies the group axioms, the natural choice being vector addition.


A homomorphism between two vector spaces $V$ and $W$ reduces to a linear map $T: V \rightarrow W$. This map is fully defined by how it trasforms the basis set $\{v_1, v_2, ...\}$, which we refer to as primitives. An ideal composition learner would preserve how parts combine - like the way vector addition in pixel space, to representation space. However, such behaviour is typically not observed in real models, primarily because of the deep nesting of non-linearities. Thus, we now focus on  \textit{learning} how to combine parts in the representation space, rather than assuming it’s just addition.

To quantify this, we study how the wavelet sub-band representations evolve through the network. In the latent space, we recompose the primitives' representations (like we do in pixel space) and compare the result with the original image. This helps reveal how compositional the learned representations are. To make this analysis manageable, we focus on compositionality in the last encoder layer.

\section{Compositionality Analysis}
\label{sec:comp_analysis}

\subsection{Drawing Parallels from Existing Works}
The inspiration for a framework to study compositionality in ViTs stems from the work by \cite{andreas2019measuring}. The paper offers a framework to measure compositionality in deep learning models, particularly neural networks. In the context of this paper, compositionality refers to the ability of a system to represent complex ideas using simpler concepts. It introduces a metric to measure how well an explicitly compositional model $\hat{f}_{\eta}$ can approximate a complex model $f$. To draw parallels, we summarize our understanding of their framework, with corresponding analogies to our approach:

 \textbf{1) Representations}: They define a model \( f: \mathcal{X} \rightarrow \Theta \), where \( \mathcal{X} \) is the input space (e.g., images), and \( \Theta \) is the representation space. The output representations \( \theta \in \Theta \) produced by $f$ are analysed for compositional behaviour.
 
\textbf{Analogy}: In our work, \( f \) refers to the ViT model, \( \mathcal{X} \) is the set of input images, and \( \Theta \) is the space of encoder representations. Each \( \theta \) represents the ViT's internal encoding of an image.

\textbf{2) Derivations}: Derivations \( \mathcal{D} \) are recursively constructed from a finite set of primitives \( \mathcal{D}_0 \) using a binary bracketing operation \( \langle \cdot, \cdot \rangle \). If \( d_i, d_j \in \mathcal{D} \), then \( \langle d_i, d_j \rangle \in \mathcal{D} \). A derivation oracle \( D: \mathcal{X} \rightarrow \mathcal{D} \) maps each input to its derivation tree.



\begin{figure}
\includegraphics[trim=3pt 8pt 3pt 6pt,width=0.9\linewidth,clip]{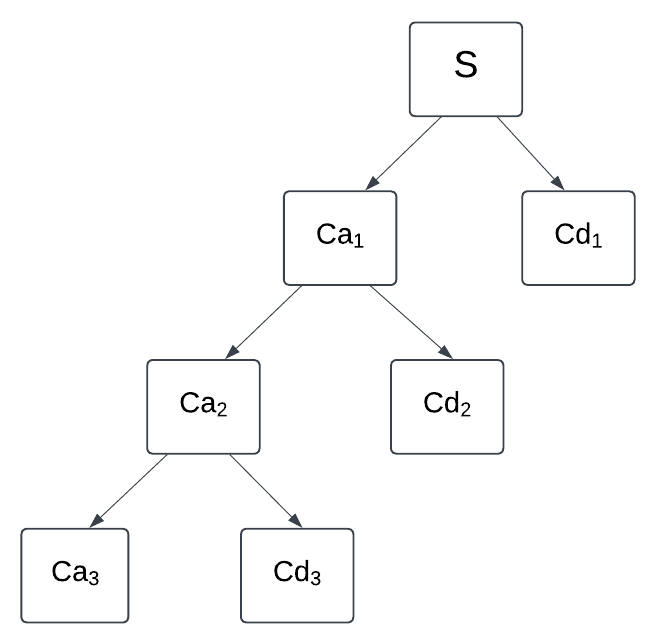} 
\caption{Tree structure of the Discrete Wavelet Transform (DWT). S represents the input signal. $Ca_{i},Cd_{i}$ represent the approximate and detail coefficients of $i^{th}$ level.}
\label{fig:wav_tree}
\end{figure} 

\textbf{Analogy:} In our framework, derivations correspond to wavelet decompositions. The DWT acts as the oracle \( D \), constructing a tree ~(Figure~\ref{fig:wav_tree}) from wavelet sub-bands. Although the set of sub-bands is infinite (\ref{subsec:comp_in_rep}), combining valid sub-bands (primitives) still yields a valid derivation.

\textbf{3) Compositionality}: The model $f$ is compositional if it preserves the structure of composition from the input space to the representation space, i.e. it is a homomorphism from input space to representation space. A composition operation on representations $*$: $\theta_{a} * \theta_{b} \mapsto \theta$ is defined such that for any input $x$ with derivation $D(x) = \langle D(x_a), D(x_b) \rangle$, we have: \begin{align*} f(x) = f(x_a) * f(x_b). \end{align*} While exactly compositional primitives may not exist, can a set of candidate primitives approximate the model's internal representation? If $f$ can be approximated by a learnable compositional model $\hat{f}_{\eta}$ with parameters $\eta$, then the approximation itself could serve as a measure of compositionality for $f$.


\textbf{Analogy}: In our case, the model $f: \mathcal{X} \longrightarrow \Theta$ is the ViT model. We consider an intermediate encoder layer $l$ such that $f_{l}: \mathcal{X} \longrightarrow \Theta$ operates from the input space to the encoding layer $l$. Then, the compositional model $\hat{f}_{\eta}(d): \mathcal{D} \longrightarrow \Theta$ can be viewed as an approximation of the encoder layer representations $f_{l}$. With this perspective, we can study the compositionality of any encoder layer of the ViT architecture.


\begin{figure*}[t!]
    \centering
    \begin{subfigure}{0.15\textwidth}
        \includegraphics[width=\textwidth]{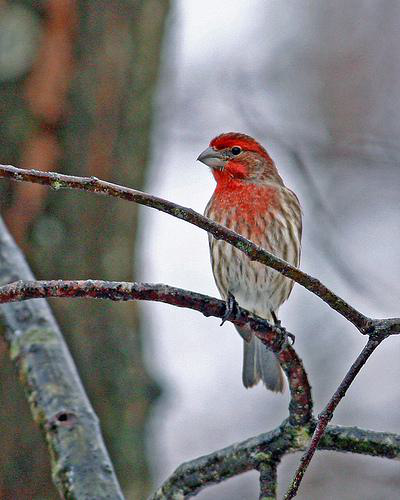}
        \caption{Image}
    \end{subfigure}
    \begin{subfigure}{0.8\textwidth}
        \includegraphics[width= 0.95\textwidth]{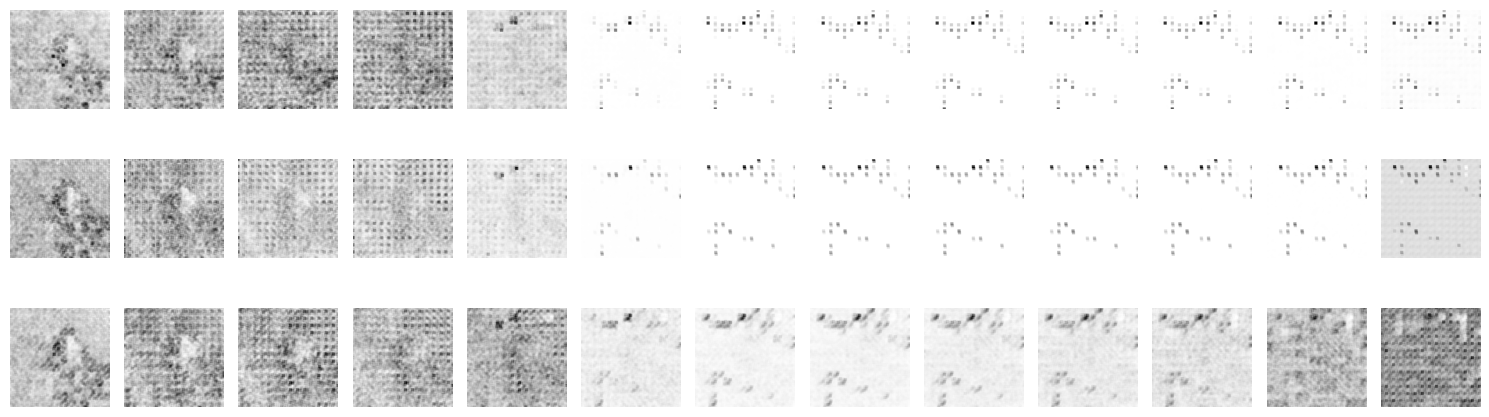}
        \caption{SSIM maps}
    \end{subfigure}
    \caption{SSIM maps for each channel (R,G,B). For each encoder layer output, the original image's representation is compared with the composed image representation. The SSIM maps shown here are \textbf{after} comparison. There is no immediate notion of compositionality present visually.}
    \label{fig:SSIM maps}
\end{figure*}

\subsection{Capturing Compositionality}
\label{subsec: Cap_com}
The wavelet reconstruction in the image space gives back the original image without any loss of information. Thus, a natural approach to check the model's compositionality would be to examine how the wavelet reconstruction behaves in the representation space of the encoder layers. We analyse if such composition of the reconstructed encoder layer representations approximates the encoder layer representations of the original image. We identify two metrics, Structural Similarity Index (SSIM)~\citep{1284395} and Centered Kernel Alignment(CKA)~\citep{kornblith2019similarityneuralnetworkrepresentations}. These metrics compare the image's encoder layer representation of the original image with that of the composed reconstruction. SSIM is a perceptual metric and takes into account local patterns of pixel intensities, their correlation, and spatial arrangements. CKA is used to compare the similarity between two sets of high-dimensional feature vectors (often from neural network layers).
Using these metrics, we conduct the below analysis:
\begin{enumerate}
    \item We use the SSIM map~\citep{1284395} to visualize structural similarities between the original and the composed representations. To do this, we reshape the encoder layer representation from $E_L(I)^{N - 1\times D}$ to $ E_L(I)^{W\times H \times C}$ where $N$ is the number of tokens ($N-1$ to exclude the \verb|CLS| token), and $D$ is the encoder layer's hidden dimension. We measure the SSIM across the channels.
    \item We plot the CKA \citep{kornblith2019similarityneuralnetworkrepresentations} scores between the image representation and composed representation across all encoder layers. For this analysis, we sample 10k images(10 images per class) from the imagenet-1k dataset and average the CKA scores over all encoder layers.
\end{enumerate}

\begin{figure}
\includegraphics[trim=3pt 8pt 3pt 6pt,width=\linewidth,clip]{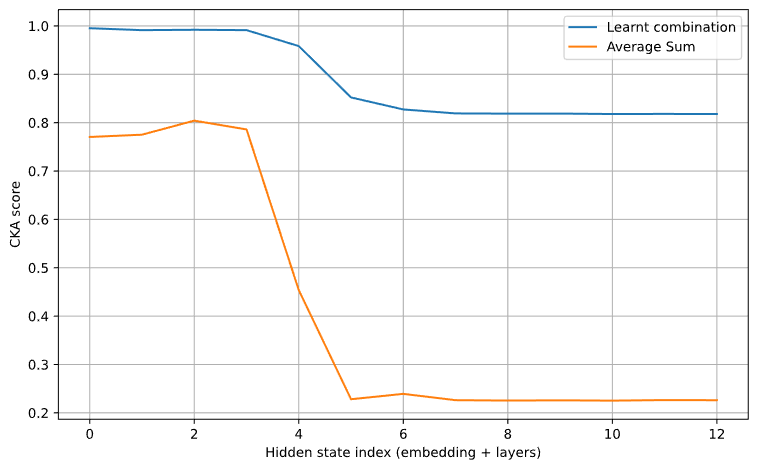} 
\caption{CKA scores of original vs composed representations at various encoder layers of ViT-B averaged over 10K images. The learned combination has a much higher CKA score across layers compared to a simple sum.}
\label{fig:cka}
\end{figure}
Figure~\ref{fig:SSIM maps} shows the SSIM maps computed for a sample image, and Figure~\ref{fig:cka} presents the CKA scores averaged over 10K images from the ILSVRC validation set using ViT-Base representations. Neither the maps nor the scores provide any evidence of compositionality or structural similarity between the representations. It is unlikely that simply adding the individual wavelet representations would exactly give the original image's representation. This invites the possibility that reconstruction of these primitives in the representation space differs from reconstruction in the image space. Motivated by this, we investigate whether such a composition function can be learned to better approximate the true representations by relaxing the constraint that each wavelet representation has to be equally weighted.

\subsection{Compositionality Framework for ViTs}
\label{subsec:Framework}
To generate a set of primitives for the pixel space, we turn to the 2D Discrete Wavelet Transform (DWT). 


Let $E_l: \mathbb{R}^{W\times H\times C}\longrightarrow \mathbb{R}^{N\times D}$ be a function that takes an input image of dimension $W\times H\times C$ and outputs a set of $N$ (number of patches + 1) token vectors of dimension $D$, produced by the $l^{\text th}$ layer of a vision transformer with $L$ encoder layers (i.e., $1 \leq l \leq L$). 

We investigate the following question to assess whether a ViT model exhibits compositionality:
\begin{equation} \label{eq:compositionality}
\sum_{l = 1}^{L} \left\| E_l(I) - \left( E_l(I_{LL}) + \sum_{m=1}^{M}\sum_{k=1}^{3} (E_lI_{\text{detail}}^{(m,k)}) \right) \right\|_2 = 0 ?
\end{equation}
Here, \( I_{LL} = \sum_{i,j} A_{M,i,j} \phi_{M,i,j} \) denotes the low-frequency (approximation) image at level \( M \), and \( I_{\text{detail}}^{(m,k)} = \sum_{i,j} D_{m,i,j}^{k} \psi_{m,i,j}^{k} \) denotes the \( k^{\text{th}} \) directional high-frequency detail (horizontal, vertical, diagonal) at level \( m \). $E_l(\cdot)$ is the ViT encoder output at layer $l$.
That is, we check if the representation of the original image at encoder layer $l$ can be reconstructed by summing up the representations of its wavelet components. The preliminary analysis presented in Figs.~\ref{fig:SSIM maps} and ~\ref{fig:cka} shows that this equality does not hold, suggesting a lack of compositionality under simple addition.

Given the highly non-linear nature of the ViT model and the high dimensionality of its representations, we relax the strict equality and ask whether a learnable function can approximate the original image's encoder layer representation:
\begin{equation} \label{eq:approximation}
E_l(I) \approx g_{\eta} \left( E_l(I_{LL}),\ \left\{ E_l\left( I_{\text{detail}}^{(m,k)} \right) \right\}_{\substack{1 \leq m \leq M \\ 1 \leq k \leq 3}} \right),
\end{equation}
where \( I_{LL} = \sum_{i,j} A_{M,i,j} \phi_{M,i,j} \) and \( I_{\text{details}}^{(m,k)} = \sum_{i,j} D_{m,i,j}^{k} \psi_{m,i,j}^{k} \).
 That is, can we approximate original representations at layer $l$ of the encoder by applying a learnable composition function $g_\eta(.)$  (with parameters $\eta$) on the primitive representations of the image? To emphasize, $g_\eta(.)$ attempts to find the best possible {\em linear combination} of the primitive representations.

We argue that popular distance metrics between these two high-dimensional representations might not be a reliable way of measuring similarity due to the curse of dimensionality. Instead, we aim to minimize the loss between the final layer classifier output logits of the original image's final CLS token and the approximate \textit{linearly combined} final layer CLS token. Since we are not modifying any of the ViT model's parameters while training this composition function, we affirm that all our analyses are post-hoc and still viable probes for understanding the pretrained ViT model. Hence, our reformulated question to evaluate whether compositionally holds becomes:
\begin{equation}
\begin{split}
    \eta^* &= \arg\min_{\eta}\\&\mathcal{L}\left(
    E_c\left(E_l(I)_{\text{[CLS]}}\right),\;
    E_c\left(g_\eta\left(\left\{E_l(\Tilde{I}_p)_{\text{[CLS]}}\right\}_{p=1}^{n}\right)\right)
    \right),
\end{split}
\end{equation}
where
$\mathcal{L}$ is the loss, $E_l(I)_{\text{[CLS]}} \in \mathbb{R}^{D}$ is the CLS token from the original image at encoder layer $l$, $\Tilde{I}_p$ denotes the $p^\text{th}$ primitive (either $\sum A_{M,i,j}\phi_{M,i,j}$ or $D_{m,i,j}^k\psi_{m,i,j}^k$), $E_l(\Tilde{I}_p)_{\text{[CLS]}}$ is its corresponding CLS token, and $g_\eta: \mathbb{R}^{n \times D} \rightarrow \mathbb{R}^{D}$ is a learnable linear function mapping $n$ primitives to a single vector.

\begin{figure*}[htpb]
    \centering    
    \includegraphics[width=0.75\textwidth,clip]{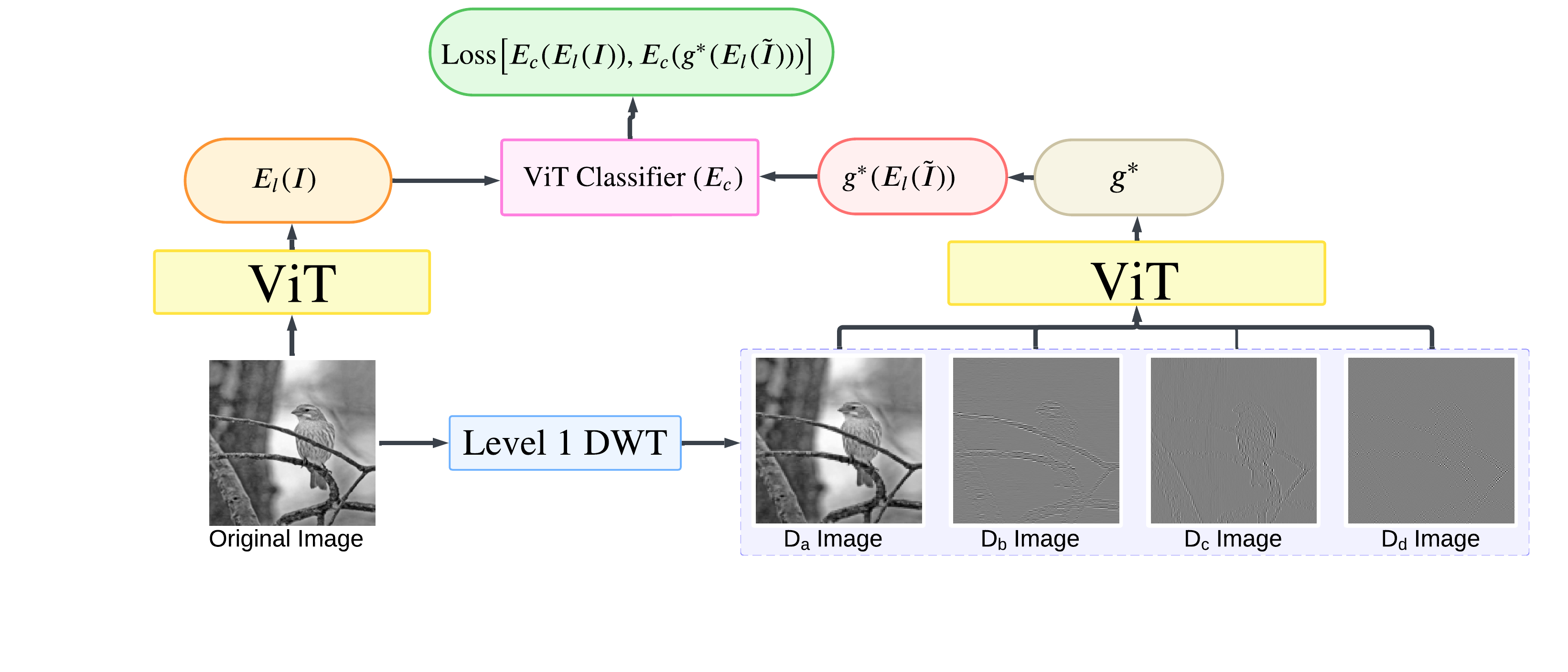}
    \caption{Overview of the proposed compositionality framework for ViTs. The figure presents learning the composition function for Level 1 DWT decomposition. $D_{a}$, $D_{b}$, $D_{c}$, $D_{d}$ are the coefficients of the wavelet decomposition discussed in \ref{eq:sum}.}
    \label{fig:Framework}
\end{figure*}

\subsection{Applications of this Framework}
Our proposed framework demonstrates that the representations at the final encoder layer exhibit compositional behaviour using DWT primitives obtained by a one-level decomposition. Now, we validate this framework's utility by evaluating it under commonly encountered real-world distortions. Specifically, we assess compositinality in the presence of additive noise and compression. Our experiments show that our compositional framework is robust and holds even in the presence of such distortions.

\section{Experimental Setup and Results}

We conduct our experiments on the ImageNet-1k dataset~\citep{5206848}, sampling 50 images per class (50,000 total) and splitting them into 60:20:20 train/val/test sets. To train the composition function \( g^* \), we first generate wavelet primitives via the DWT and pass them through the ViT to obtain their encoder layer outputs \( E_l(\Tilde{I}) \). The CLS tokens from these outputs serve as inputs to \( g^* \), which outputs a composed CLS token. This is fed to the ViT classifier, and the target is the original image’s classification output—not the image label—as our aim is to measure how well the composed token reproduces the original representation. Only \( g^* \) is trained (using cross-entropy loss), while ViT weights remain frozen. Models are trained for 100 epochs using SGD with a learning rate of 0.001.

We restrict our analysis to two levels of DWT decomposition using two wavelet bases: Haar and db4. To assess generalizability, we evaluate two ViT variants—ViT-B and ViT-L—both pretrained on ImageNet-21k (14M images, 21k classes). We relax the equal-weight constraint from Section~\ref{subsec: Cap_com} and explore three adaptive weighting strategies for the composition function \( g^* \):
\begin{enumerate}
    \item \textbf{Convex:} weights satisfy \( \sum_i \eta_i = 1 \), \( \eta_i \geq 0 \);
    \item \textbf{Conic:} non-negative weights \( \eta_i \geq 0 \);
    \item \textbf{Unconstrained:} no restrictions on \( \eta \).
\end{enumerate}

We used the same subset of images from the ImageNet-1k to learn the composition function $g^*$ for these three variations. While the framework can be used to study any encoder layer in the model, we restrict our analysis to the last layer,  whose outputs are often inputs to downstream tasks. 

        


\subsection{Composition Approximation: Accuracy of Learned Model on Ground Truth}

Our initial analysis brings us back to our first question (eq.~\ref{eq:compositionality}) posed in section~\ref{subsec:Framework}: whether wavelet based primitive representations, satisfy compositionality under simple summation. We compare the classification accuracy of the representations composed following simple summation (eq.~\ref{eq:compositionality}) and that of the learned composition model. Table~\ref{tab:accuracies_comparison} compares the classification accuracies for three cases  (i) original ViT's output, (ii) output from summing the individual wavelet decomposition representations, and (iii) output from the proposed learned composition model. Please note, these accuracies are calculated on the ground truth. These results clearly demonstrate that the learned representations outperform the summed representations significantly. Notably, the performance of level 1 decomposition is almost on par with that of ViT model's accuracy. 
These findings confirm that the learned composition function $g^*$ is effective in capturing the compositionality of level 1 wavelet primitives.
\begin{table*}[h!]
    \centering
    \scalebox{0.9}{
    \begin{tabular}{|c|c|c|c|c|c|}
    \hline
         &  &  & \multicolumn{3}{c|}{Learned} \\ \hline
        Model &Original&Summed & Unconstrained& Conic& Convex \\ \hline
        ViT-B (Haar-level 1) &0.792 &0.13 &0.775 &0.775 &0.771 \\ \hline
        ViT-B (db4-level 1) &0.792 &0.13 & 0.777& 0.775& 0.772\\ \hline
        ViT-L (Haar-level 1) &0.809 &0.18 &0.797 &0.795 &0.795 \\ \hline
        ViT-B (Haar-level 2) &0.83&0.005 &0.51 &0.5 &0.48 \\ \hline
        ViT-B (db4-level 2) & 0.83& 0.005 & 0.51&0.51 &0.48 \\ \hline
        ViT-L (Haar-level 2) &0.82 &0.003 &0.63 &0.62 &0.59 \\ \hline
    \end{tabular}}
    \caption{Accuracies of original representations vs. summed representations vs. learned compositions. Note that the learned representations perform significantly better
than just the summed representations.}
    \label{tab:accuracies_comparison}
\end{table*}

\begin{table}[h!]
    \centering
    \scalebox{0.9}{
    \begin{tabular}{|c|c|c|c|}
    \hline
        Model & Unconstrained&Conic&Convex \\ \hline
        ViT-B (haar-level 1) &0.87 &0.87& 0.86\\ \hline
        ViT-B (db4-level 1) & 0.9& 0.9& 0.89 \\ \hline
        ViT-L (haar-level 1) &0.92 &0.91 &0.91 \\ \hline
        ViT-B (haar-level 2) &0.53 &0.51 &0.49 \\ \hline
        ViT-B (db4-level 2) &0.69 &0.68 &0.61  \\ \hline
        ViT-L (haar-level 2) &0.65 &0.64 &0.61  \\ \hline
    \end{tabular}}
    \caption{Relative accuracy of the learned composition models. Note that the target for the composed representation is the output predicted by the original image classifier (not the ground truth label).}
    \label{tab:Relative Accs}
\end{table}



\subsection{Composition Approximation: Understanding the Learned Model Weights}

\begin{table*}[h!]
    \small  
    \centering
    \begin{tabular}{|c|c|c|c|}
    \hline
         Model& Unconstrained & Conic & Convex \\ \hline
         ViT-B (haar) & [ 2.02, -0.18,  0.43,  0.18] & [1.67, 0.34, 0.57,  0.02] & [0.66, 0.11, 0.10, 0.12] \\ \hline
         ViT-B (db4)  & [ 2.02, 0.1,  -0.15, -0.16] & [1.65, 0.12, 0.63, 0.03] & [0.62, 0.09, 0.25, 0.03] \\ \hline 
         ViT-L (haar) & [ 1.93, 0.16,  -0.02,  0.25] & [1.81, 0.28, 0.13, 0.44] & [0.68, 0.1, 0.05, 0.16] \\ \hline
    \end{tabular}
    \caption{Weights learned by the proposed composition model $(g^*)$ for level 1 wavelet decomposition.}
    \label{tab:Weights_level_2}
\end{table*}
\begin{table*}[h!]    
    \centering
    \scalebox{0.8}
    {
    \begin{tabular}{|c|c|c|c|}
    \hline
         Model& Unconstrained & Conic & Convex \\ \hline
         ViT-B (haar) & [1.32,  0.35, -0.07, -0.14,  0.65, -0.20,  0.21] & [1.88, 0.61, 0.35, 0.17, 0.10, 0.10, 0.44] & [0.42, 0.13, 0.05, 0.13, 0.07, 0.10, 0.06] \\ \hline
         ViT-B (db4)  & [1.52, -0.18,  0.06,  0.30,  0.35,  0.16, -0.21] & [1.64, 0.40, 0.12, 0.02, 0, 0.03, 0] & [0.43, 0.11, 0.08, 0.07, 0.14, 0.06, 0.08] \\ \hline 
         ViT-L (haar) & [1.52, -0.01, -0.21,  0.29,  0.06, -0.01,  0.34] & [1.82, 0.29, 0.32, 0.17, 0, 0.33, 0.23] & [0.40, 0.11, 0.10, 0.08, 0.09, 0.06, 0.13] \\ \hline
    \end{tabular}
    }
    \caption{Weights learned by the proposed composition model $(g^*)$ for level 2 wavelet decomposition.}
    \label{tab:Weights_level_1}
\end{table*}
To evaluate how accurately our learned composition function $g^{*}$ approximates the original image's representation, we compute the relative accuracy (by considering the \textbf{original model's (ViT)} output logits as the ground truth, or reference target). Table~\ref{tab:Relative Accs} presents those results. Interestingly, the relative accuracies are similar across different constraints (convex, conic, and unconstrained variations of $g^*$). To investigate this further, we look at the learned model weights in Table~\ref{tab:Weights_level_2} and Table~\ref{tab:Weights_level_1}, which indicate the relative importance weights assigned to different sub-bands.  Across all settings, the learned model $g^*$ weighs the approximation (low-pass filtered image) coefficient i.e, the first value significantly more than the other coefficients in the representation space. 

Notably, there is no discernible pattern among the learned weights under different constraints. There is considerable variation among the weights assigned for different $g^*$'s, but their performance is quite similar. There could be multiple such compositions for an encoder layer, which leads to further questions about the representation space. 

\subsection{Composition Approximation: Learned Reconstructed Image Analysis}
\begin{figure*}[ht!]
\centering
\begin{subfigure}{0.7\textwidth}
    \includegraphics[width=\textwidth]{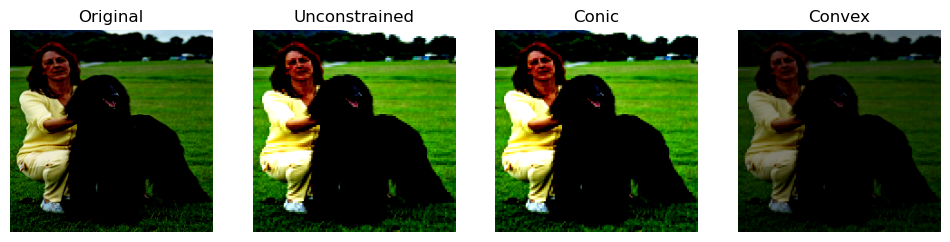}
    \caption{Reconstructed Images for Level 1 ViT-B (Haar) model}
    \label{fig:fourth_1}
\end{subfigure}
\begin{subfigure}{0.7\textwidth}
    \includegraphics[width=\textwidth]{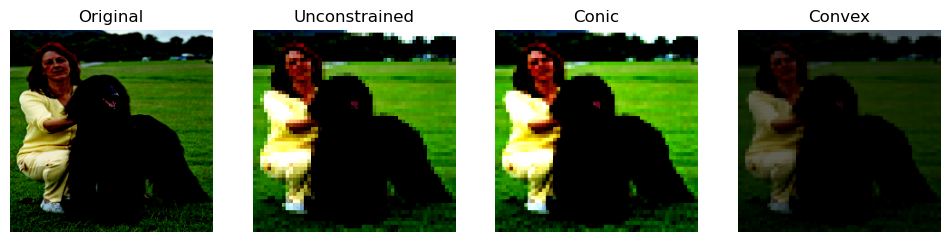}
    \caption{Reconstructed Images for Level 2 ViT-B (Haar) model}
    \label{fig:fourth_2} 
\end{subfigure}
\caption{Reconstructed images obtained by applying the learned composition model $g^*$'s weights to the corresponding sub-bands of the original image in the input space.}
\label{fig:reconstruct}
\end{figure*}

In this subsection, we investigate how the weights learned by the proposed composition function $(g^*)$ influence the reconstruction of the original images, when these weights are applied to the primitives (wavelet sub-bands) in the image space. Simply put, we transfer the weights learned from the ViT encoder embedding space to their corresponding wavelet sub-bands in the image space, thereby reconstructing a weighted image in the image space. We consider a subset of 200 images for this analysis. Table~\ref{tab:my_label} presents the classification accuracy of the ViT model on the reconstructed images.  Note that although there is a significant drop in level 2 accuracies, the learned weights translate back well for level 1 decomposition. Figure~\ref{fig:reconstruct} visualizes the reconstructed images using Level 1 ViT-B (Haar) and Level 2 ViT-B (Haar) model. Interestingly, although the convex combination of the sub-bands in the image space significantly affect the pixel intensities, their performance is at par with other learned models.

\begin{table}[h]
    \centering
    \scalebox{0.9}{
    \begin{tabular}{|c|c|c|c|c|}
    \hline
        Model &Original& Unconst.& Conic&Convex \\ \hline
        ViT-B (haar-level 1) & 0.79& 0.72 & 0.72&0.76\\ \hline
        ViT-B (db4-level 1)  & 0.84&0.81& 0.81 & 0.82 \\ \hline
        ViT-B (haar-level 2) &0.84&0.58 &0.48 &0.64 \\ \hline
        ViT-B (db4-level 2) &0.84&0.49 &0.51 &0.65 \\ \hline
        ViT-L (haar-level 1) &0.83& 0.82 & 0.82 & 0.80 \\ \hline
        ViT-L (haar-level 2) &0.83&0.63 & 0.68&0.71  \\ \hline
    \end{tabular}}
    \caption{Classification accuracy of ViT-B on reconstructed images generated by applying the learned weights of the proposed composition model $(g^*)$ to the corresponding sub-bands in image space.}
    \label{tab:my_label}
\end{table}

\subsection{Composition Approximation: Error Analysis}
\label{subsec:error-analysis}
While the classification performance of the proposed compositional model presented thus far provides a broad picture of compositionality, a natural question about error in composition arises. Here,  
we compare the compositional model's predictions - particularly its misclassification - with that of the original model. Since our downstream task is image classification, we analyze the composition error via prediction discrepancies. It would be interesting to explore other ways to study the error in composition in the future.
In this preliminary experiment, we sample $1000$ images from the Imagenet-1K dataset, and the performance of both the original and compositional model (level 1 DWT decomposition) is evaluated as follows:
\begin{itemize}
    \item Percentage of images where the original model is accurate and the learned model is inaccurate (Err\textsubscript{Learned$\neg$Org}). 
    \item Percentage of images where the learned model is accurate and the original model is inaccurate (Err\textsubscript{Org$\neg$Learned}).
    \item Percentage of images where both models are inaccurate (Err\textsubscript{both}).
\end{itemize}
\begin{table*}[htpb]
    \centering
    \scalebox{0.9}{
    \begin{tabular}{|c|c|c|c|c|c|}
        \hline        Model&Err\textsubscript{Learned}&Err\textsubscript{Org}&Err\textsubscript{Learned$\neg$Org}&Err\textsubscript{Org$\neg$Learned}&Err\textsubscript{both}\\ \hline
         ViT-B\textsubscript{Unconstrained}(Level-1 haar) & 19.7\% & 17.1\% & 3.8\% & 1.2\% & 15.9\%\\ \hline
         ViT-B\textsubscript{Conic}(Level-1 haar) & 19.7\% & 17.1\% & 3.8\%& 1.2\% & 15.9\%\\ \hline
         ViT-B\textsubscript{Convex}(Level-1 haar) & 20.4\%&17.1\% &4.3\% & 1\%& 16.1\%\\ \hline
    \end{tabular}}
    \caption{We report the errors on the test set using the learned composition model. The reported percentages are calculated on the 1000-image subset. Interestingly, there is a fraction of samples on which the learned composition performs better than the original model.}
    \label{tab:err}
\end{table*}
The results in Table~\ref{tab:err} provides a comparative analysis of prediction error between the learned model and the original ViT. As anticipated, the learned model commits relatively more errors than the original. However, it is noteworthy that the learned model also performs better on some images. This preliminary analysis provides sufficient motivation to further analyse the role of individual wavelet representations towards the model's prediction. 

\subsection{Composition Approximation: Effect of Distortion on Images}
To further cement the practical utility of our proposed framework, we explore whether it persists when images are subjected to distortions. We measure how JPEG compression and additive Gaussian noise affect the classification accuracy of our learned model.
\begin{table*}[htbp]
    \centering
    \scalebox{0.9}{
    \begin{tabular}{|c|c|ccc|}
    \hline
    \multirow{2}{*}{\centering Image Type} & \multirow{2}{*}{\centering Image Accuracy} & \multicolumn{3}{c|}{Learned Accuracy} \\ \cline{3-5}
    & & Unconstrained & Conic & Convex \\ \hline
    Original Images   & 0.792 & 0.775 & 0.775 & 0.771 \\ \hline
    Compressed Images & 0.628 & 0.603 & 0.603 & 0.599 \\ \hline
    Noisy Images      & 0.593 & 0.565 & 0.565 & 0.563 \\ \hline
    \end{tabular}}
    \caption{Comparison of classification accuracies for original, compressed, and noisy images. Learned accuracies are obtained from the output of the composition model \( g^* \).}
    \label{tab:subband_accuracy}
\end{table*}
Table~\ref{tab:subband_accuracy} clearly demonstrates that our framework is robust to distortions, and compositoniality holds even with compressed and noisy images.

\section{Conclusion and Future Work}
Our work explores notions of compositionality present in the ViT encoder layer representations. We present a general framework to measure compositional behaviour in encoder layers of ViT-based architectures. Fundamental to this framework is the use of the DWT representation as an input-dependent primitive. Our findings indicate the possibility of compositional behaviour in the ViT model. Specifically, we provide evidence for compositionality in the last encoder layer when primitives induced by a one-level DWT decomposition are applied. While our present analysis is restricted to the final encoder layer, we aim to explore all the encoder layers for potential compositionality. We hope this work leads to further analysis for explainability in ViT's.

{
    \small
    \bibliographystyle{ieeenat_fullname}
    \bibliography{ref}
}

\end{document}

%% file: ref.bib
@article{daubechies1992ten,
  title={Ten lectures on wavelets},
  author={Daubechies, Ingrid},
  journal={Society for industrial and applied mathematics},
  year={1992}
}

@InProceedings{radford2021learning,
  title = 	 {Learning Transferable Visual Models From Natural Language Supervision},
  author =       {Radford, Alec and Kim, Jong Wook and Hallacy, Chris and Ramesh, Aditya and Goh, Gabriel and Agarwal, Sandhini and Sastry, Girish and Askell, Amanda and Mishkin, Pamela and Clark, Jack and Krueger, Gretchen and Sutskever, Ilya},
  booktitle = 	 {Proceedings of the 38th International Conference on Machine Learning},
  pages = 	 {8748--8763},
  year = 	 {2021},
  editor = 	 {Meila, Marina and Zhang, Tong},
  volume = 	 {139},
  series = 	 {Proceedings of Machine Learning Research},
  month = 	 {18--24 Jul},
  publisher =    {PMLR},
  url = 	 {https://proceedings.mlr.press/v139/radford21a.html}
}

@inproceedings{
park2022vision,
title={How Do Vision Transformers Work?},
author={Namuk Park and Songkuk Kim},
booktitle={International Conference on Learning Representations},
year={2022},
url={https://openreview.net/forum?id=D78Go4hVcxO}
}

@inproceedings{bhojanapalli2021understanding,
  title={Understanding robustness of transformers for image classification},
  author={Bhojanapalli, Srinadh and Chakrabarti, Ayan and Glasner, Daniel and Li, Daliang and Unterthiner, Thomas and Veit, Andreas},
  booktitle={Proceedings of the IEEE/CVF international conference on computer vision},
  pages={10231--10241},
  year={2021}
}

@inproceedings{vaswani2023attention,
 author = {Vaswani, Ashish and Shazeer, Noam and Parmar, Niki and Uszkoreit, Jakob and Jones, Llion and Gomez, Aidan N and Kaiser, \L ukasz and Polosukhin, Illia},
 booktitle = {Advances in Neural Information Processing Systems},
 editor = {I. Guyon and U. Von Luxburg and S. Bengio and H. Wallach and R. Fergus and S. Vishwanathan and R. Garnett},
 pages = {},
 publisher = {Curran Associates, Inc.},
 title = {Attention is All you Need},
 url = {https://proceedings.neurips.cc/paper_files/paper/2017/file/3f5ee243547dee91fbd053c1c4a845aa-Paper.pdf},
 volume = {30},
 year = {2017}
}

@inproceedings{
dosovitskiy2020image,
title={An Image is Worth 16x16 Words: Transformers for Image Recognition at Scale},
author={Alexey Dosovitskiy and Lucas Beyer and Alexander Kolesnikov and Dirk Weissenborn and Xiaohua Zhai and Thomas Unterthiner and Mostafa Dehghani and Matthias Minderer and Georg Heigold and Sylvain Gelly and Jakob Uszkoreit and Neil Houlsby},
booktitle={International Conference on Learning Representations},
year={2021},
url={https://openreview.net/forum?id=YicbFdNTTy}
}

@inproceedings{caron2021emerging,
  title={Emerging properties in self-supervised vision transformers},
  author={Caron, Mathilde and Touvron, Hugo and Misra, Ishan and J{\'e}gou, Herv{\'e} and Mairal, Julien and Bojanowski, Piotr and Joulin, Armand},
  booktitle={Proceedings of the IEEE/CVF international conference on computer vision},
  pages={9650--9660},
  year={2021}
}

@inproceedings{he2022masked,
  title={Masked autoencoders are scalable vision learners},
  author={He, Kaiming and Chen, Xinlei and Xie, Saining and Li, Yanghao and Doll{\'a}r, Piotr and Girshick, Ross},
  booktitle={Proceedings of the IEEE/CVF conference on computer vision and pattern recognition},
  pages={16000--16009},
  year={2022}
}

@inproceedings{li2022exploring,
  title={Exploring plain vision transformer backbones for object detection},
  author={Li, Yanghao and Mao, Hanzi and Girshick, Ross and He, Kaiming},
  booktitle={European Conference on Computer Vision},
  pages={280--296},
  year={2022},
  organization={Springer}
}

@inproceedings{strudel2021segmenter,
  title={Segmenter: Transformer for semantic segmentation},
  author={Strudel, Robin and Garcia, Ricardo and Laptev, Ivan and Schmid, Cordelia},
  booktitle={Proceedings of the IEEE/CVF international conference on computer vision},
  pages={7262--7272},
  year={2021}
}

@inproceedings{
raghu2022vision,
title={Do Vision Transformers See Like Convolutional Neural Networks?},
author={Maithra Raghu and Thomas Unterthiner and Simon Kornblith and Chiyuan Zhang and Alexey Dosovitskiy},
booktitle={Advances in Neural Information Processing Systems},
editor={A. Beygelzimer and Y. Dauphin and P. Liang and J. Wortman Vaughan},
year={2021},
url={https://openreview.net/forum?id=R-616EWWKF5}
}

@article{FODOR19883,
title = {Connectionism and cognitive architecture: A critical analysis},
journal = {Cognition},
volume = {28},
number = {1},
pages = {3-71},
year = {1988},
issn = {0010-0277},
doi = {https://doi.org/10.1016/0010-0277(88)90031-5},
url = {https://www.sciencedirect.com/science/article/pii/0010027788900315},
author = {Jerry A. Fodor and Zenon W. Pylyshyn},
}

@article{Chalmers1990WhyFA,
author="Chalmers, D.",
title="Why Fodor and Pylyshyn were wrong : the simplest refutation",
journal="Proceedings of the Twelfth Annual Conference of the Cognitive Science Society, Cambridge",
year="1990",
pages="340-347",
URL="https://cir.nii.ac.jp/crid/1570854174742444672"
}

@inproceedings{andreas2019measuring,
  author       = {Jacob Andreas},
  title        = {Measuring Compositionality in Representation Learning},
  booktitle    = {7th International Conference on Learning Representations, {ICLR} 2019,
                  New Orleans, LA, USA, May 6-9, 2019},
  publisher    = {OpenReview.net},
  year         = {2019},
  url          = {https://openreview.net/forum?id=HJz05o0qK7},
  bibsource    = {dblp computer science bibliography, https://dblp.org}
}

@article{janfreg2001,
 ISSN = {09258531, 15729583},
 URL = {http://www.jstor.org/stable/40180264},
 author = {Theo M. V. Janssen},
 journal = {Journal of Logic, Language, and Information},
 number = {1},
 pages = {115--136},
 publisher = {Springer},
 title = {Frege, Contextuality and Compositionality},
 urldate = {2024-01-21},
 volume = {10},
 year = {2001}
}

@article{Chen2023SkillsinContextPU,
  title={Skills-in-Context Prompting: Unlocking Compositionality in Large Language Models},
  author={Jiaao Chen and Xiaoman Pan and Dian Yu and Kaiqiang Song and Xiaoyang Wang and Dong Yu and Jianshu Chen},
  journal={ArXiv},
  year={2023},
  volume={abs/2308.00304},
  url={https://api.semanticscholar.org/CorpusID:260351132}
}

@inproceedings{Li2023DissectingCC,
  title={Dissecting Chain-of-Thought: Compositionality through In-Context Filtering and Learning},
  author={Yingcong Li and Kartik K. Sreenivasan and Angeliki Giannou and Dimitris Papailiopoulos and Samet Oymak},
  booktitle={Neural Information Processing Systems},
  year={2023},
  url={https://api.semanticscholar.org/CorpusID:265051253}
}

@inproceedings{
Dziri2023FaithAF,
title={Faith and Fate: Limits of Transformers on Compositionality},
author={Nouha Dziri and Ximing Lu and Melanie Sclar and Xiang Lorraine Li and Liwei Jiang and Bill Yuchen Lin and Sean Welleck and Peter West and Chandra Bhagavatula and Ronan Le Bras and Jena D. Hwang and Soumya Sanyal and Xiang Ren and Allyson Ettinger and Zaid Harchaoui and Yejin Choi},
booktitle={Thirty-seventh Conference on Neural Information Processing Systems},
year={2023},
url={https://openreview.net/forum?id=Fkckkr3ya8}
}

@INPROCEEDINGS{5206848,
  author={Deng, Jia and Dong, Wei and Socher, Richard and Li, Li-Jia and Kai Li and Li Fei-Fei},
  booktitle={2009 IEEE Conference on Computer Vision and Pattern Recognition}, 
  title={ImageNet: A large-scale hierarchical image database}, 
  year={2009},
  volume={},
  number={},
  pages={248-255},
  doi={10.1109/CVPR.2009.5206848}
}

@ARTICLE{1284395,
  author={Zhou Wang and Bovik, A.C. and Sheikh, H.R. and Simoncelli, E.P.},
  journal={IEEE Transactions on Image Processing}, 
  title={Image quality assessment: from error visibility to structural similarity}, 
  year={2004},
  volume={13},
  number={4},
  pages={600-612},
  doi={10.1109/TIP.2003.819861}
}

@misc{yao2022wavevitunifyingwavelettransformers,
      title={Wave-ViT: Unifying Wavelet and Transformers for Visual Representation Learning}, 
      author={Ting Yao and Yingwei Pan and Yehao Li and Chong-Wah Ngo and Tao Mei},
      year={2022},
      eprint={2207.04978},
      archivePrefix={arXiv},
      primaryClass={cs.CV},
      url={https://arxiv.org/abs/2207.04978}, 
}

@misc{zhang2024waveletformernettransformerbasedwaveletnetwork,
      title={WaveletFormerNet: A Transformer-based Wavelet Network for Real-world Non-homogeneous and Dense Fog Removal}, 
      author={Shengli Zhang and Zhiyong Tao and Sen Lin},
      year={2024},
      eprint={2401.04550},
      archivePrefix={arXiv},
      primaryClass={cs.CV},
      url={https://arxiv.org/abs/2401.04550}, 
}

@misc{kornblith2019similarityneuralnetworkrepresentations,
      title={Similarity of Neural Network Representations Revisited}, 
      author={Simon Kornblith and Mohammad Norouzi and Honglak Lee and Geoffrey Hinton},
      year={2019},
      eprint={1905.00414},
      archivePrefix={arXiv},
      primaryClass={cs.LG},
      url={https://arxiv.org/abs/1905.00414}, 
}
